\documentclass{ecai} 

\usepackage{latexsym}
\usepackage{amssymb}
\usepackage{amsmath}
\usepackage{amsthm}
\usepackage{booktabs}
\usepackage{enumitem}
\usepackage{graphicx}
\usepackage{color}

\newcommand{\BibTeX}{B\kern-.05em{\sc i\kern-.025em b}\kern-.08em\TeX}

\begin{document}


\begin{frontmatter}


\paperid{1820} 


\title{Integrating Expert Knowledge into \\Logical Programs via LLMs}


\author[A]{\fnms{Franciszek}~\snm{Górski}
\thanks{Corresponding Author. Email: franciszek.gorski@pg.edu.pl}\footnote{Equal contribution.}}
\author[B, D]{\fnms{Oskar}~\snm{Wysocki}
\footnotemark}
\author[D]{\fnms{Marco}~\snm{Valentino}
}
\author[B,C,D]{\fnms{Andre}~\snm{Freitas}
}

\address[A]{Multimedia Systems Department, Gdansk University of Technology, Poland}
\address[B]{Department of Computer Science, University of Manchester, United Kingdom}
\address[C]{National Biomarker Centre (NBC), CRUK Manchester Institute, United Kingdom}
\address[D]{Idiap Research Institute, Martigny, Switzerland}


\begin{abstract}
This paper introduces \textbf{ExKLoP}, a novel framework designed to evaluate how effectively Large Language Models (LLMs) integrate expert knowledge into logical reasoning systems.
This capability is especially valuable in engineering, where expert knowledge—such as manufacturer-recommended operational ranges—can be directly embedded into automated monitoring systems. By mirroring expert verification steps, tasks like range checking and constraint validation help ensure system safety and reliability. Our approach systematically evaluates LLM-generated logical rules, assessing both syntactic fluency and logical correctness in these critical validation tasks.
We also explore the models' capacity for self-correction via an iterative feedback loop based on code execution outcomes. ExKLoP presents an extensible dataset comprising 130 engineering premises, 950 prompts, and corresponding validation points. It enables comprehensive benchmarking while allowing control over task complexity and scalability of experiments. We leverage the synthetic data creation methodology to conduct extensive empirical evaluation on a diverse set of LLMs including Llama3, Gemma3, Codestral and QwenCoder. The results reveal that most models generate nearly perfect syntactically correct code and exhibit strong performance in translating expert knowledge into correct code. At the same time, while most LLMs produce nearly flawless syntactic output, their ability to correctly implement logical rules varies, as does their capacity for self-improvement. Overall, ExKLoP serves as a robust evaluation platform that streamlines the selection of effective models for self-correcting systems while clearly delineating the types of errors encountered.
\end{abstract}
\end{frontmatter}


\section{Introduction}

Large Language Models (LLMs) have demonstrated remarkable abilities in translating natural language into executable code, opening new possibilities for integrating expert knowledge into Expert Systems. This capability is particularly promising in domains such as engineering, where technical manuals, guidelines, and service reports could serve as inputs for building expert-driven reasoning systems. Such systems would enable a new class of knowledge-informed machine learning models, seamlessly combining data-driven approaches with domain expertise.

However, for this vision to become a reality, we must first assess how effectively LLMs can translate domain-specific facts and logical rules into executable code. For instance, consider an industrial machine with multiple operational parameters and manufacturer-recommended ranges for normal operation. Ideally, an LLM should be able to encode these recommendations into a program that can automatically monitor the machine’s performance. Yet, the accuracy, reliability, and robustness of such model-generated code remain open questions.

To address these challenges, we require well-defined frameworks, evaluation datasets, and benchmarks to systematically assess LLMs' ability to generate executable reasoning rules. Expert systems demand precise logical rules, making it essential to investigate whether LLMs, even when initially incorrect, can self-correct (also via agentic systems) by leveraging feedback from external symbolic models (e.g., a Python interpreter) \cite{jiang2024leanreasoner, pan2023logic, kalyanpur2024llm, bi2024iterative, xu2024faithful}. The envisioned future of systems capable of producing fully functional reasoning modules through iterative refinement hinges on key architectural decisions—such as model selection, prompt formulation, and feedback mechanism design. 

In this paper, we introduce \textbf{ExKLoP}, a framework designed to assess LLMs’ capability to integrate expert knowledge into logical reasoning systems while evaluating their potential for self-correction. Our framework provides a structured approach to exploring the reliability of LLM-generated code via syntax, runtime and logic validation, before and after self-correction based on the output errors. We offer insights into both model performance and iterative improvement strategies.

To guide our investigation, we formulate the following key research questions:\newline
\textbf{RQ1}: Can large language models (LLMs) facilitate the integration of expert knowledge with AI and knowledge-informed machine learning models in engineering?\newline
\textbf{RQ2}: How efficiently can LLMs convert expert knowledge expressed in natural language into valid Python logic rules?\newline
\textbf{RQ3}: To what extent can LLMs self-correct their mistakes using feedback from external program interpreter?

To address these questions, we present the following contributions:\newline
$\bullet$ We introduce \textbf{ExKLoP}, an experimental framework for evaluating how effectively LLMs integrate expert knowledge from engineering domain into logical rules and executable code by measuring Formalization Success Rate and Logical Consistency Rate. Additionally, our framework assesses the models' capacity for self-correction by leveraging feedback from the execution environment.\newline
$\bullet$ ExKLoP provides a simple yet powerful approach to exploring the potential of agentic systems in knowledge-based reasoning. Specifically, we investigate whether iterative self-correction enhances the accuracy and reliability of model-generated outputs.\newline
$\bullet$ We introduce a scalable and extensible dataset generation methodology for engineering premises and validation points. ExKLoP, comprising 130 premises and 950 prompts, enables efficient benchmarking while allowing researchers to control task complexity and seamlessly expand experiments by integrating additional premises within a single prompt.\newline
$\bullet$ We demonstrate significant performance variations in generating correct logical rules depending on the size and type of the LLM, whether it is a general-purpose model or one specialized in code generation, as well as on the difficulty of the task. While all LLMs produce nearly flawless syntactic output, their ability to correctly implement logical rules varies, as does their capacity for self-improvement.

\section{ExKLoP}
\begin{figure*}[h!]
    \centering
  \includegraphics[width=.85\textwidth]{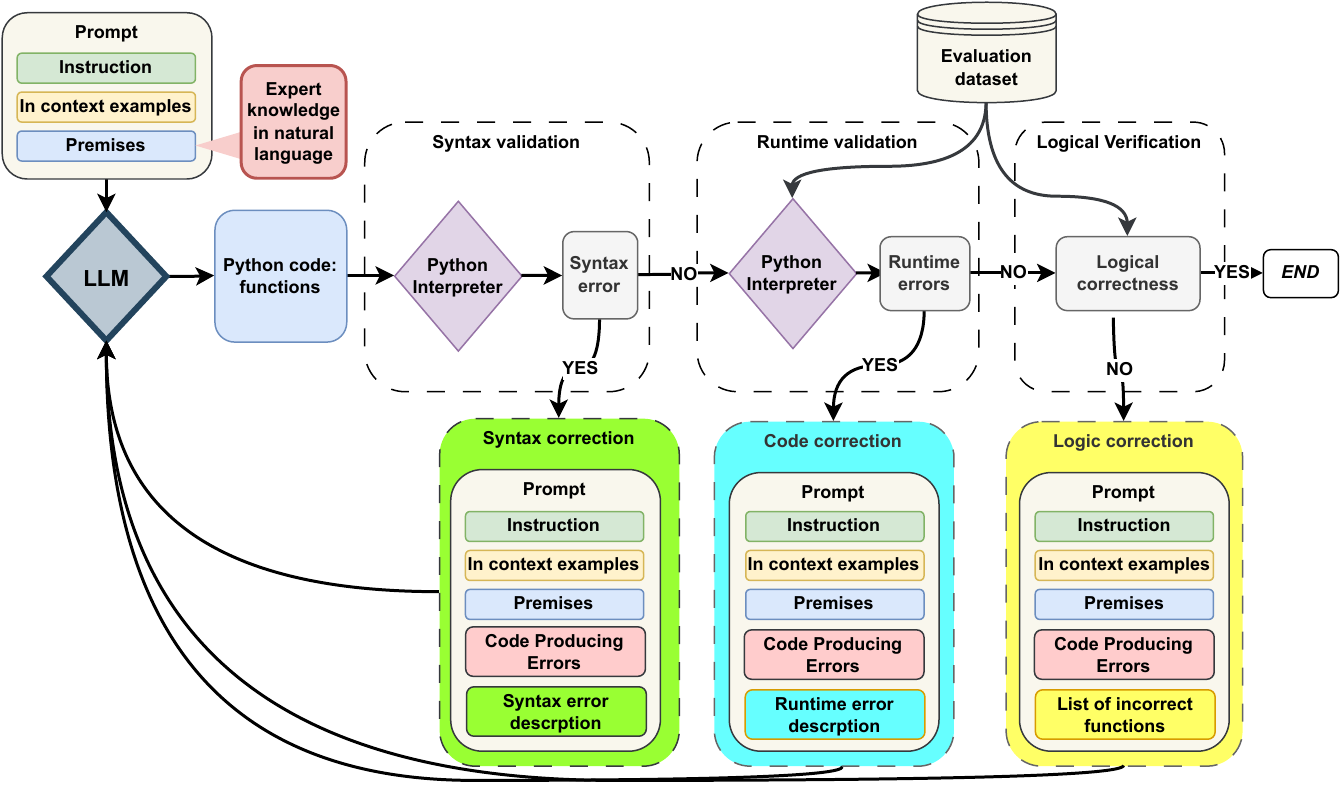}
  \caption {ExKLoP framework for evaluating expert knowledge integration in Python-based logical rules}
  \label{fig:python_rules_gen_diagram}
\end{figure*}

ExKLoP follows a structured evaluation process consisting of three key phases: (1) Translation of natural language premises into executable functions, (2) Syntax, runtime, and logic validation, and (3) Iterative refinement of incorrect outputs using an LLM-driven self-correction loop. This end-to-end workflow is illustrated in Figure \ref{fig:python_rules_gen_diagram}.

\subsection{Translation of Natural Language to Python Logic}
The primary task assigned to the LLM is to generate functions that accurately encode logical constraints specified in natural language. These premises define acceptable ranges for given physical parameters of an object, ensuring compliance with expected operational conditions. In engineering applications, such constraints often describe the normal operating ranges of mechanical systems. For example, an industrial machine may have predefined thresholds for temperature, pressure, and vibration that indicate safe operation. The prompt contains instruction, in-context examples and premises to be translated into code (for details see Section \ref{prompt_construction}).

\subsection{Syntax, runtime, and logic validation}
\begin{figure}[t]
\centering
  \includegraphics[width=.7\linewidth]{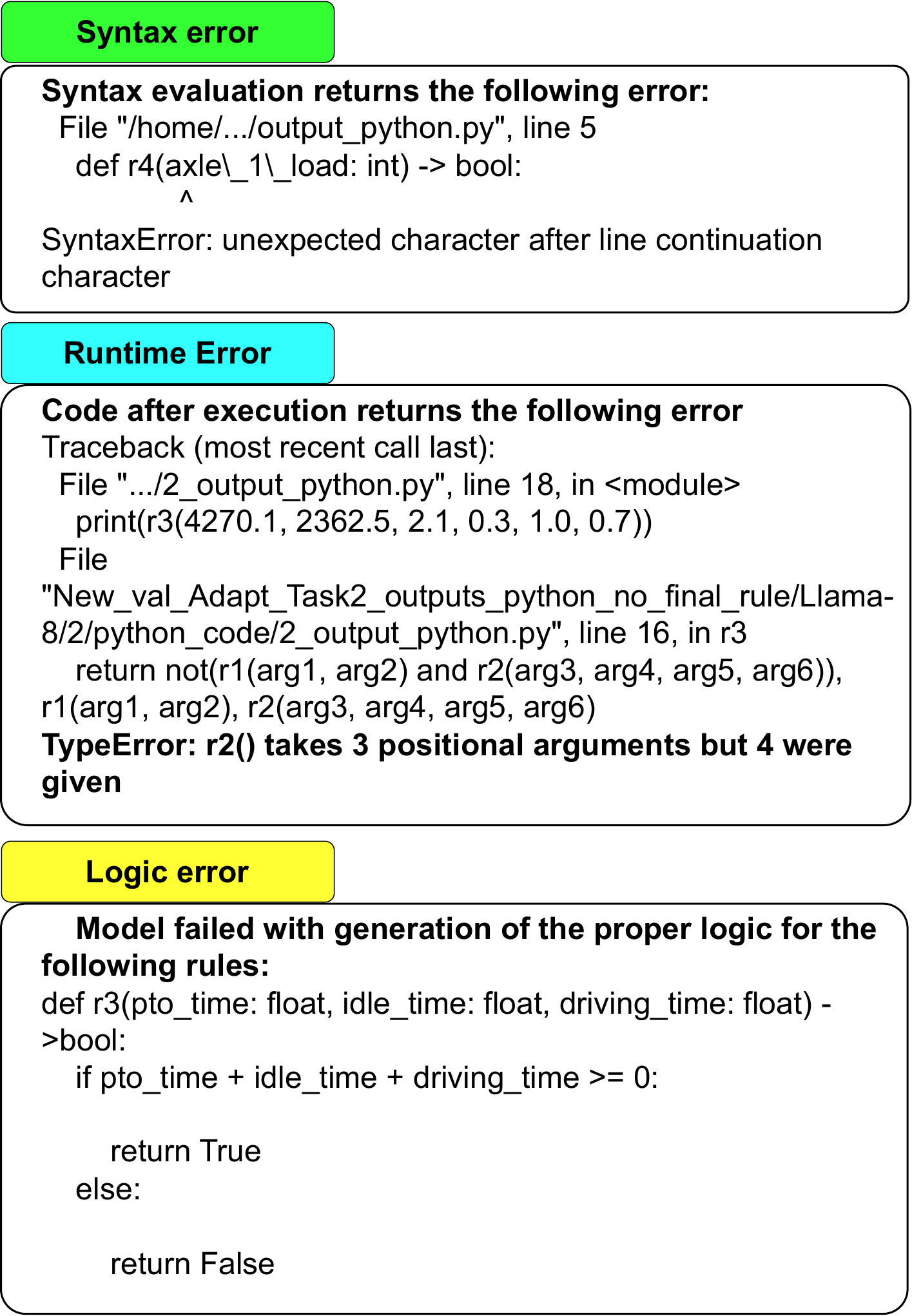} 
  \caption {Examples of syntax, runtime and logic errors that are used in the self-correction. }
  \label{fig:Errors_examples}
\end{figure}

Once the Python functions are generated, they undergo a multi-stage validation process to ensure correctness across three dimensions:
\newline
\textbf{Syntax Validation:} The Python interpreter checks for syntactical correctness detecting any parsing errors, such as missing colons or incorrect indentation.
\newline
\textbf{Runtime Validation:} The generated code is executed using the validation dataset as a source of input data points to identify runtime errors, such as mismatched argument counts or undefined variables. 
\newline
\textbf{Logical Verification}: This step determines whether the logic implemented by the model accurately represents the intended constraints. The functions take data points from the validation dataset and evaluate their outputs against ground truth labels.
\newline
Any errors are flagged for correction and used as added part of the prompt in the following step (see examples in Fig.\ref{fig:Errors_examples}).

\subsection{Iterative Refinement via LLM Self-Correction}
If a function fails any validation step, it undergoes an iterative refinement process. This involves re-prompting the LLM with structured feedback to improve the function. Each re-prompt includes a slightly modified task instruction, the previously generated code, and a detailed error description to guide correction. The refinement process consists of three key stages:

\textbf{Syntax Correction:} Syntax errors are provided to the LLM with explicit error messages detailing the cause of failure. The model is then prompted to regenerate the function while ensuring that the identified syntax issue is corrected.
\newline
\textbf{Execution Error Correction:} For runtime errors, the corresponding error message is provided as feedback to the LLM for examples such as an incorrect number of arguments or undefined variables). The model then attempts to revise the function to ensure successful execution.
\newline
\textbf{Logical Correction:} For incorrect outputs during logical verification, the LLM is tasked with refining the conditional logic. It receives feedback on the specific functions that are part of the overall logic, allowing it to correct these functions accordingly.

After each refinement step, the updated function is re-evaluated using the same validation process (see Fig. \ref{fig:python_rules_gen_diagram}). The refinement loop continues for up to 10 steps or until no further improvements are observed.

\subsection{Metrics}
The framework uses two metrics: 

\textbf{Formalization Success Rate (FSR)} measures the proportion of natural language inputs correctly translated into syntax error-free Python rules, assessing the LLM's ability to generate valid code.
\begin{equation}
    \label{eq:fa}
    FSR = \frac{X_f}{X}
\end{equation}


\textbf{Logical Consistency Rate (LCR)} measures the proportion of natural language inputs accurately translated into Python rules that not only execute without errors but also correctly determine whether a data point is an outlier, ensuring logical validity.
\begin{equation}
    \label{eq:oa}
    LCR = \frac{X_c}{X}
\end{equation}

where:  \(X\) is the total set of natural language inputs (prompts) containing rules; \(X_f\) is the subset of \(X\) that is successfully formalized into valid, syntax error-free Python rules; \(X_c\) is the subset of \(X_f\) where the generated rules correctly identify out-of-range values.

\subsection{Tasks}
\textbf{Task 1: Range Checking} The initial task aims to verify whether a single parameter's value lies within a predefined range, e.g., `A vehicle operates for between 2 and 10 hours per day'. 
Additionally, a single prompt contains multiple premises, each defining different parameter ranges. As the number of premises increases, so does the complexity of the task, requiring the model to correctly handle multiple constraints simultaneously. The objective is to evaluate the model’s ability to process and encode multiple premises within a single prompt, rather than just a single constraint.  

Formally, given multiple parameters $x$ with their respective valid ranges \([x_{\min}, x_{\max}]\), the generated function should determine whether all parameters satisfy their constraints:
\small
\[
F(x_1, x_2, \dots, x_n) =
\begin{cases} 
1, & \text{if } \forall i \in \{1, \dots, n\}, \\ & \quad x_i \in [x_{i,\min}, x_{i,\max}] \\
0, & \text{otherwise}
\end{cases}
\]
\normalsize

\textbf{Task 2: Constraint Validation} 
The second task extends the complexity by introducing interdependencies between parameters. Instead of evaluating each parameter independently, the constraints now define relational conditions between them. For example, in a vehicle load distribution scenario, a constraint may state: `The load on the first axle cannot be greater than the load on the second axle.'
Formally, let \( R \) be a set of relational constraints defined over the parameters:
\small
\[
R = \{ (x_i, x_j) \mid x_i \ \mathcal{O}_{ij} \ x_j, \quad \forall (i, j) \}
\]
\normalsize

where \( \mathcal{O}_{ij} \) represents a relational operator such as \( \leq, \geq \). The function \( F(x_1, x_2, \dots, x_n) \) should evaluate whether all given constraints hold simultaneously:
\small
\[
F(x_1, x_2, \dots, x_n) =
\begin{cases} 
1, & \text{if } \forall (x_i, x_j) \in R, \quad x_i \ \mathcal{O}_{ij} \ x_j \\
0, & \text{otherwise}
\end{cases}
\]
\normalsize

\subsection{Premises Dataset}
The dataset consists of statements defining the normal operating ranges of various parameters based on industry standards. To enhance linguistic diversity, each statement is rephrased into five distinct textual variations while preserving logical consistency. This approach increases robustness and ensures variability in how constraints are expressed. Each premise is crafted to resemble expert descriptions, providing realistic inputs for LLMs.

The dataset is easily extendable, with the current version covering: 17 parameters corresponding to real-world vehicle operational variables such as speed, distance, fuel consumption, and axle loads; 130 engineering premises, 950 prompts, and corresponding validation points.

\subsection{Prompt Construction}
\label{prompt_construction}
\begin{figure*}
  \centering
  \includegraphics[width=.9\textwidth]{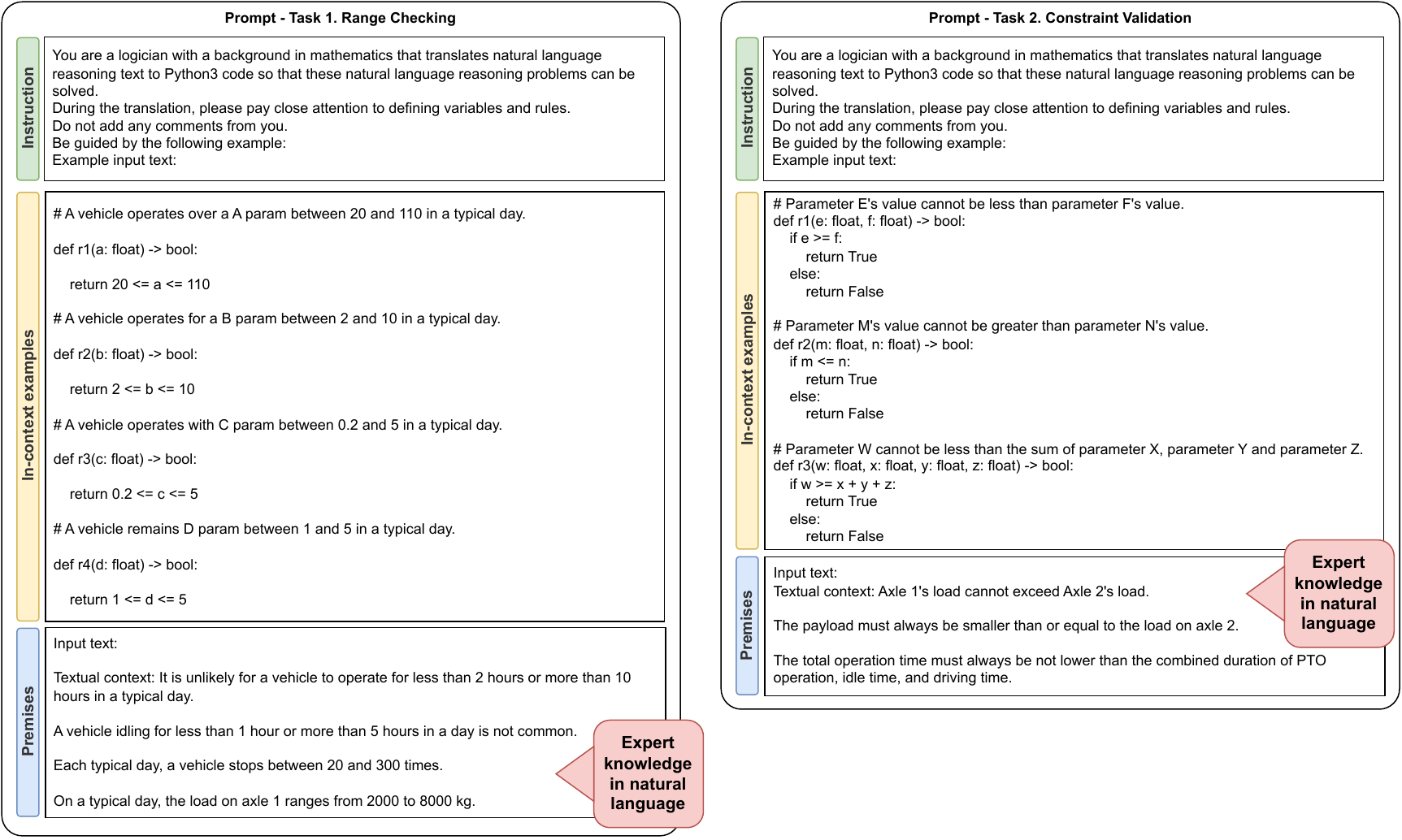} 
  \caption {Examples of prompts in Task 1. Range Checking (left) and Task 2. Constraint Validation (right).}
  \label{fig:prompt_structure}
\end{figure*}

Each prompt follows a structured format to guide the LLM in code generation and iterative refinement (Fig. \ref{fig:prompt_structure}):\newline
\textbf{Task Instruction:} A system message that clearly defines the model’s role and objective in translating natural language constraints into executable logic. \newline
\textbf{In-Context Examples:} Demonstrations of correctly formatted Python functions, with parameter names and units replaced by placeholders. This ensures the model understands the expected output format while preventing memorization of specific values. The number of examples corresponds to the number of premises in the task. \newline
\textbf{Input Premises:} A set of natural language statements defining constraints on the operational parameters of an object. These premises serve as the basis for generating Python functions and are derived from \textit{Premises Dataset}. \newline
\textbf{Error Messages (During Self-Correction):} If the generated function fails syntax, runtime, or logic validation, error messages (Fig.\ref{fig:Errors_examples}) are appended to the prompt. These messages provide explicit feedback on the failure, enabling the model to iteratively refine and correct its output.

As illustrated in Figure \ref{fig:python_rules_gen_diagram} and \ref{fig:prompt_structure}, the prompt is dynamically augmented during refinement steps, allowing the model to improve incorrect functions based on structured feedback. This iterative approach ensures better alignment between the generated code and the intended logical constraints.

\subsection{Evaluation Data Points}
As part of the ExKLoP framework, we generate a structured evaluation dataset to systematically assess whether LLM-generated functions correctly classify in-range and out-of-range values. This dataset is tailored to each task, ensuring comprehensive validation of the model’s ability to enforce predefined constraints.

\textbf{Range Checking Task:} 
For each parameter \(x_i\) (\(i=1,\dots,n\)), we generate four test points: $\{x_{i1}, x_{i2}, x_{i3}, x_{i4}\}$,with the following properties:
\[
\begin{aligned}
x_{i1} &< x_{i,\min}, \\
x_{i2},\, x_{i3} &\in [x_{i,\min}, x_{i,\max}], \\
x_{i4} &> x_{i,\max}.
\end{aligned}
\]

To minimize the total number of evaluations—which would otherwise require testing all \(n!\) combinations—we adopt the following strategy that requires only \(4n\) evaluations: for a given parameter \(x_i\), we evaluate \(F\) by varying \(x_i\) over its four test points while keeping all other parameters fixed at a nominal value \(x_{j0}\) (with \(x_{j0} \in [x_{j,\min}, x_{j,\max}]\) for \(j\neq i\)). In this way, the vector of evaluation points for parameter \(x_i\) is constructed as:
\[
\mathbf{x}^{(i,\ell)} = \bigl( x_{10}, \dots, x_{i-1,0}, x_{i\ell}, x_{i+1,0}, \dots, x_{n0} \bigr)
\]
where $\ell=1,2,3,4$. By design, for each test vector \(\mathbf{x}^{(i,\ell)}\) the function evaluation simplifies to:
\small
\[
F\bigl(\mathbf{x}^{(i,\ell)}\bigr) =
\begin{cases}
1, & \text{if } x_{i\ell} \in [x_{i,\min}, x_{i,\max}], \\
0, & \text{if } x_{i\ell} \notin [x_{i,\min}, x_{i,\max}],
\end{cases}
\]
\normalsize
since \(F(x_{j0})=1\) for all \(j\neq i\).

\textbf{Constraint Validation Task}: 
For each relational constraint \(x_i \,\mathcal{O}_{ij}\, x_j\) (with \(\mathcal{O}_{ij}\) a relational operator such as \(\geq\) or \(\leq\)), we generate three test points for the parameter \(x_i\): $\{x_{i1},\, x_{i2},\, x_{i3}\}$ with the following properties:
$x_{i1},\, x_{i2}$ satisfy $x_i \,\mathcal{O}_{ij}\, x_{j,0}$ and $x_{i3}$ violates $x_i \,\mathcal{O}_{ij}\, x_{j,0}$,
where \(x_{j,0}\) is a nominal valid value for \(x_j\) (i.e., one that complies with its range and any applicable constraints).

For a given constraint \(x_i \,\mathcal{O}_{ij}\, x_j\), we evaluate \(F\) by varying \(x_i\) over its three test points while keeping all other parameters fixed at nominal values \(x_{k,0}\) (with \(x_{k,0}\) chosen so that \(F\) evaluates to 1 when no constraint is violated). In this way, the evaluation vector is constructed as:
\[
\mathbf{x}^{(i,\ell)} = \bigl( x_{10}, \dots, x_{i-1,0}, x_{i\ell}, x_{i+1,0}, \dots, x_{n0} \bigr)
\]
where $\ell=1,2,3$. 
By design, for each test vector \(\mathbf{x}^{(i,\ell)}\) the function evaluation is:
\small
\[
F\bigl(\mathbf{x}^{(i,\ell)}\bigr) =
\begin{cases}
1, & \text{if } x_{i\ell} \,\mathcal{O}_{ij}\, x_{j,0} \quad (\ell=1,2), \\
0, & \text{if } x_{i\ell} \notin \{\,x \mid x \,\mathcal{O}_{ij}\, x_{j,0}\,\} \quad (\ell=3),
\end{cases}
\]
\normalsize
since the nominal values for the other parameters ensure that all other constraints are satisfied.

Assuming each constraint is associated with a distinct parameter being varied, the total number of evaluation scenarios for \(F(x_1, x_2, \dots, x_n)\) is \(3n\).

\section{Experiments}
To evaluate the effectiveness of the proposed ExKLoP, we conducted experiments on two tasks: Range Checking and Constraint Validation.

For the Range Checking task, each prompt comprises between 2 and 12 premises. We generated 50 prompts for each set size, resulting in a total of 550 unique prompts. Varying the number of premises per prompt allows us to assess whether performance degrades as the input complexity increases. Each prompt is assigned a unique, randomly selected set of premises to ensure diversity in parameter selection.

For the Constraint Validation task, each prompt contains between 2 and 9 conditions. With 50 prompts generated for each set size, this task includes a total of 400 unique prompts. Each prompt randomly selects a distinct set of conditions, ensuring a broad range of logical interdependencies.

In total, we evaluated 6650 prompts, as each of the 7 LLMs was tested on 550 Range Checking prompts and 400 Constraint Validation prompts.

\subsection{Experimental setting}

To ensure a comprehensive evaluation, we tested 7 open-sourced LLMs: \textbf{Llama3-70B}, \textbf{Llama3-8B} \citep{grattafiori2024llama3herdmodels}, \textbf{Gemma3-12B} \citep{team2025gemma}, \textbf{Codestral-22B} \citep{mistral2024codestral} and 3 size variants of \textbf{QwenCoder2.5} \citep{hui2024qwen2} model - \textbf{QwenCoder2.5-1.5B}, \textbf{QwenCoder2.5-3B} and \textbf{QwenCoder2.5-14B} and we refer to this models as \textbf{QwenCoderSmall}, \textbf{QwenCoderMedium} and \textbf{QwenCoder}. All models are used in their instruction-tuned (Instruct) versions, optimized for task-following. The first-step inference is performed with a temperature of 0.0 and no sampling, minimizing randomness in responses. In contrast, the refinement steps are executed with a temperature of 0.6, top\_p=0.9, top\_k=50, and sampling enabled to introduce more variety in the models’ behavior during self-improvement. Self-correction was not used in every case. It was never applied for the Llama3-70B model, by virtue of the fact that it was the largest model, with the highest computational costs, and whose results were so satisfactory that self-correction was not needed. In addition, in Task 1, models whose results were above 90\% were not subjected to correction due to the relatively low complexity of the task, requiring no additional computational expense for such results.

All experiments were conducted using 4 NVIDIA L40 GPUs, each with 48GB of VRAM, and an AMD EPYC 75F3 32-Core Processor. The downloaded models required 360GB of free disk space. All scripts were executed with Python 3.10. The pretrained LLM weights were utilized through Hugging Face’s \textit{transformers} library (version 4.33.1) and \textit{accelerate} (version 0.33.0), with \textit{CUDA} 12.1. The models were loaded using the \textit{AutoModelForCausalLM} and \textit{AutoTokenizer} classes with the options \textit{device\_map="auto"} and \textit{torch\_dtype=torch.bfloat16}, enabling multi-GPU inference and reduced GPU memory usage. 
The refinement process continues for up to 10 steps or until no further improvements are observed.

\section{Results and discussion}
Our evaluation of LLM-generated Python functions for range checking and constraint validation tasks revealed several key findings.

\textbf{High Fluency in Syntactically Correct Code}:
Most models achieved near-perfect Formalization Success Rates (only LLama3-8B has 0.99), demonstrating a strong capability in producing syntactically correct Python code. Although generating code that passes syntax checks is relatively straightforward, ensuring logical correctness is considerably more challenging. This highlights the need for robust evaluation frameworks to systematically assess not only syntax but also the underlying logic.

\textbf{Translation of Expert Knowledge into Code}:
Most models demonstrated strong performance in translating expert knowledge into code, with relatively few issues. The most common problems were runtime errors, particularly in constraint validation tasks, such as mismatched function arguments or undefined variables. Another frequent issue was logical inconsistency - for instance, reversing the order of arguments (e.g., generating X2 <= X1 instead of X1 >= X2). Additional errors included incorrect argument definitions or the omission of key physical parameters.

\begin{figure}[t]
    \centering

  \includegraphics[width=0.4\textwidth]{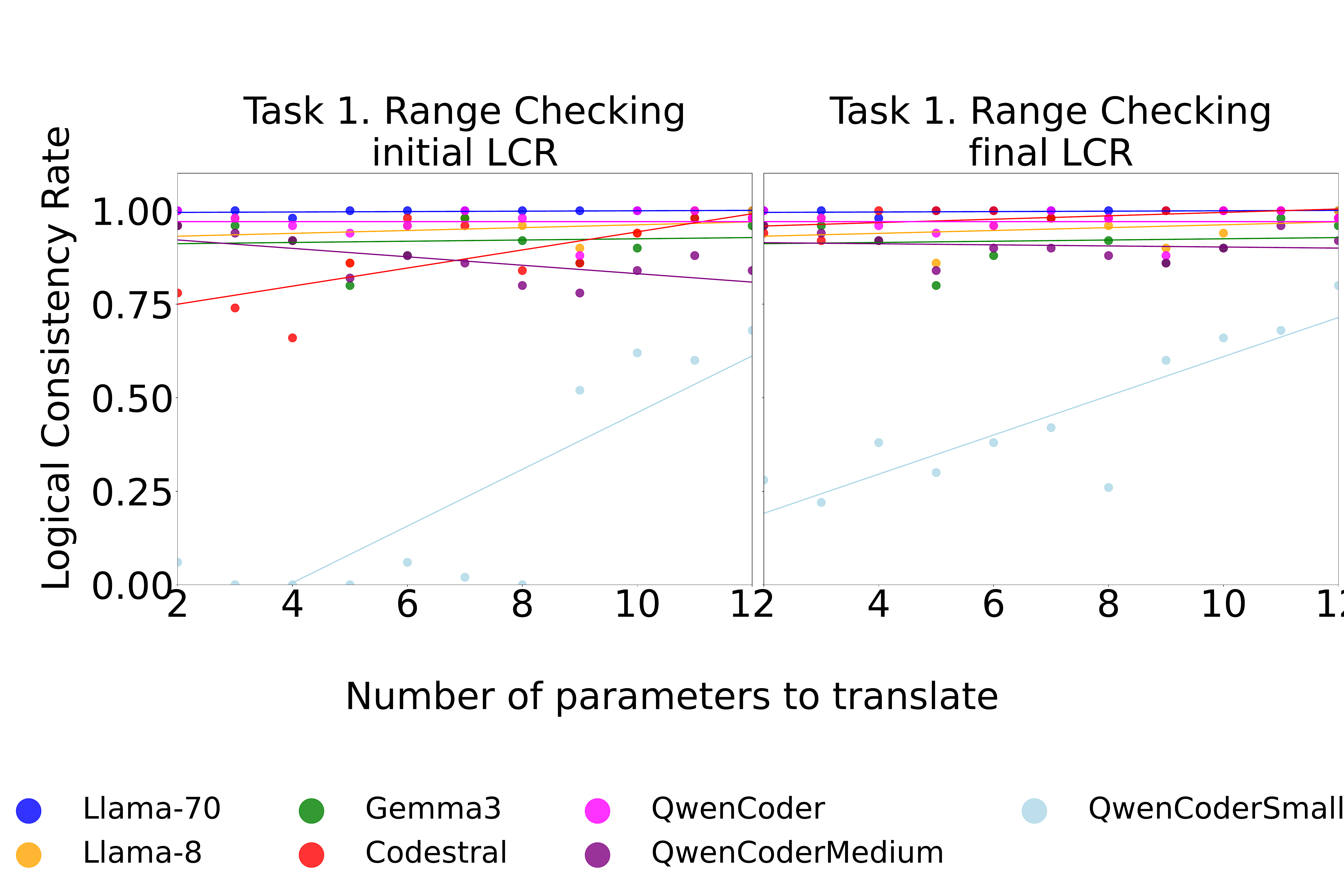}
  \caption {\textbf{Range Checking:} Logical Consistency Rate relation to task complexity, before (left) and after (right) all self-corrections.}
  \label{fig:task1_OA_before_after_comp}
\end{figure}

\begin{figure}[t]
    \centering

  \includegraphics[width=\columnwidth]{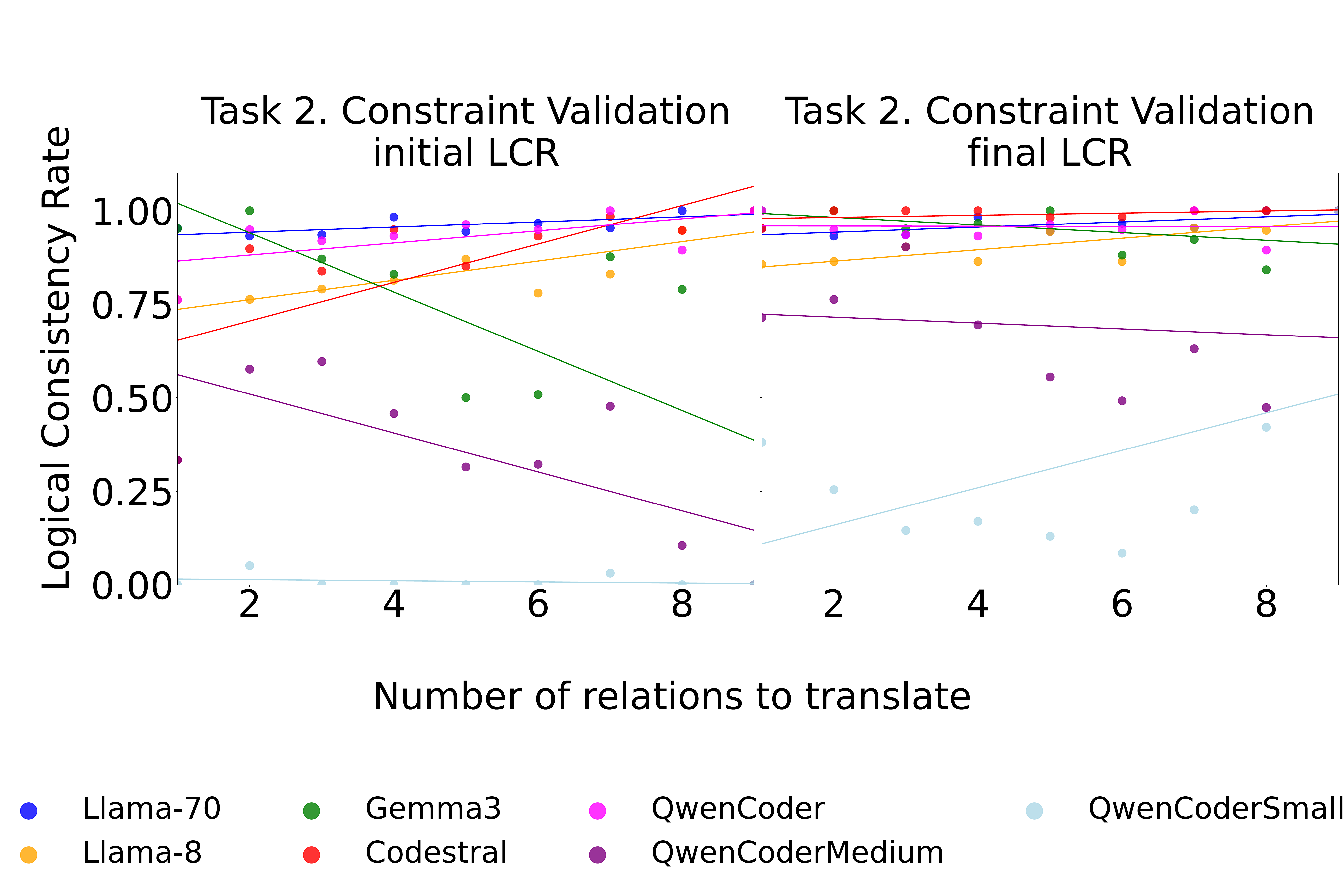}
  \caption {\textbf{Constraint Validation:} Logical Consistency Rate relation to task complexity, before (left) and after (right) all self-corrections.}
  \label{fig:task2_OA_before_after_comp}
\end{figure}

\textbf{Impact of the number of parameters/relations in a prompt}:
As we can in see in Figures \ref{fig:task1_OA_before_after_comp} and \ref{fig:task2_OA_before_after_comp} in both tasks, the relationship between the number of input parameters (or logical relationships) and the Logical Correctness Rate (LCR) proved to be somewhat unintuitive. In the Range Checking task, increasing the number of parameters in the input consistently improved performance, even for models with otherwise low overall LCR scores. This suggests that, for the models studied, a higher number of input parameters was not a hindrance—rather, it appeared to facilitate better reasoning. In contrast, this trend was partially reversed in the Constraint Validation task. Some models—such as Gemma3, QwenCoderMedium, and QwenCoderSmall—performed worse with more parameters. However, other models maintained the same positive correlation observed in the Range Checking task.

\textbf{Task 1: Logical Correctness in Range Checking}:
most models achieved very high results with Llama3 models and QwenCoder being the best. Only the smallest QwenCoderSmall model has very poor performance, under 25\% and stops his self-refinement process after 6 steps with result nearly doubled burt still low, only 45\%. Interestingly, Codestral scored lower than smaller models, just 87\%, and need 5 refinement steps to achieve 98\% Logical Correcntness Score. Noteworthy is the performance of the QwenCoderMedium model, which initially reached 87\%, the same as Codestral, which is 7 times larger, and improved by an additional 4 percentage points over 4 steps.


\textbf{Task 2: Logical Correctness in Constraint Validation}:
the largest model, Llama3-70B, led with a Logical Correctness Rate (LCR) of 0.96, followed by QwenCoder at 0.94, Codestral at 0.88, and Llama3-8B at 0.82. After two steps of self-refinement, Codestral improved its score to 0.99, achieving the highest result overall. Llama3-8B improved to 0.90 after two steps, QwenCoder reached 0.95 after a single step, and Gemma3 made a notable leap from a modest initial LCR of 0.78 to 0.95 in just two refinement steps. QwenCoderMedium started with a much lower score of 44\% and increased it to 67\% over six refinement steps, representing the largest relative improvement across both tasks. QwenCoderSmall began with a very low LCR of 0.02 and reached 0.19 after six steps of refinement.

\begin{figure}[t]
    \centering

  \includegraphics[width=\columnwidth]{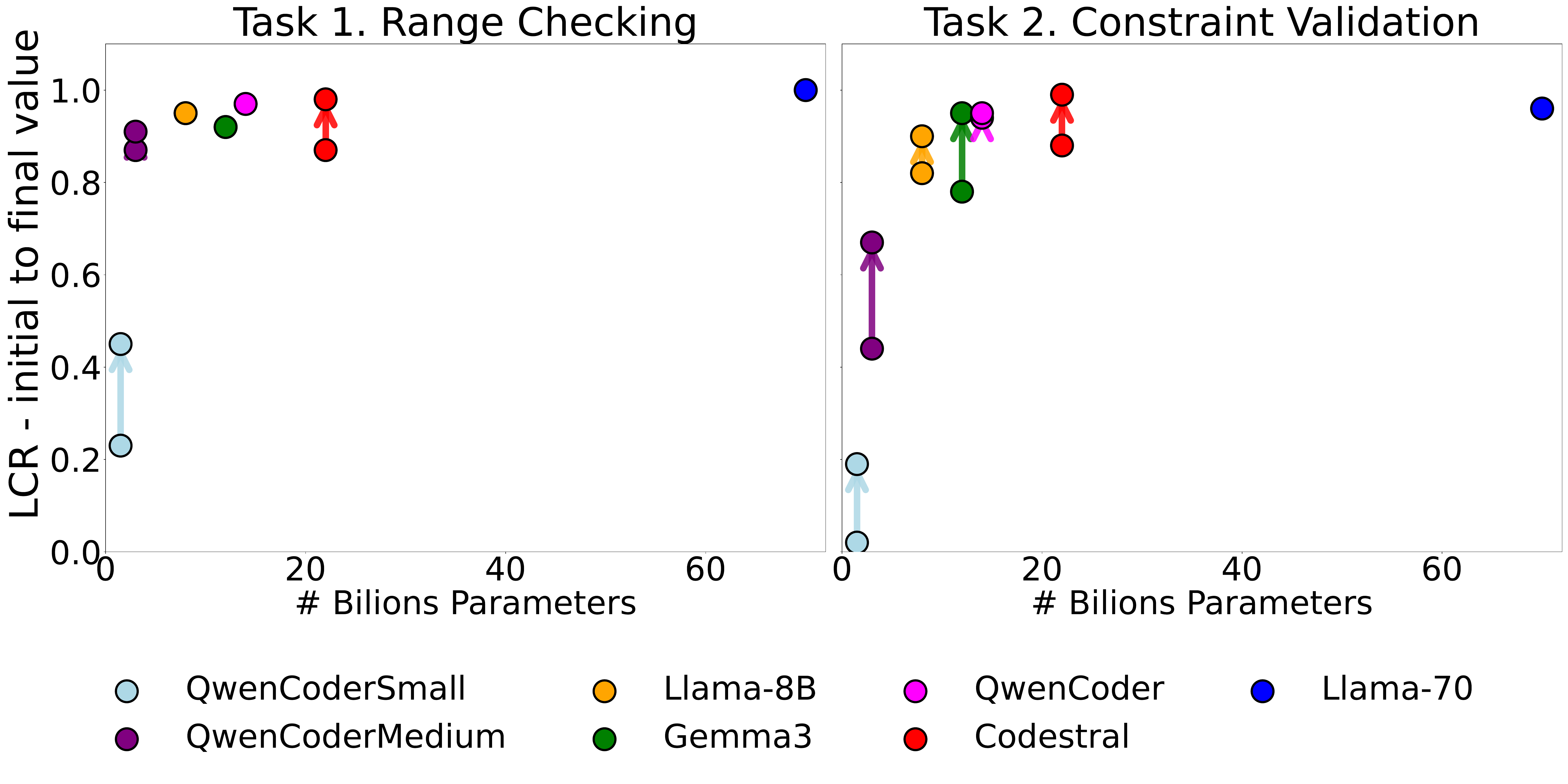}
  \caption {Logical Consistency Rate vs. Model Size. Arrows indicate changes from the initial to final values after improvement iterations}
  \label{fig:lcr_vs_models}
\end{figure}

\textbf{Impact of Self-Correction and Model Comparison}:
Self-correction proved to be a key factor influencing model performance. However, the extent of its impact varied depending on the model and the task type. All models benefited from iterative self-correction, but larger models (Codestral, Gemma3, Llama3-8B) achieved meaningful improvements in fewer steps compared to smaller models (QwenCoderMedium, QwenCoderSmall). Notably, in none of the tested cases did self-correction proceed more than 6 steps, typically halting earlier. Increasing the variability in model responses by adjusting the temperature and enabling sampling positively influenced the effectiveness of the self-correction process.
Figure \ref{fig:lcr_vs_models} indicates that a larger number of parameters does not necessarily lead to significantly better performance. A model like Llama3-70B appears oversized for optimal use in both tasks, as comparable results can be achieved by much smaller models—such as Llama3-8B, Gemma3, or QwenCoder—with just a few additional inference steps. While Llama3-70B may reach high performance in a single inference, this comes at a substantially higher computational cost.

\begin{figure}[t]
    \centering
  \includegraphics[width=\columnwidth]{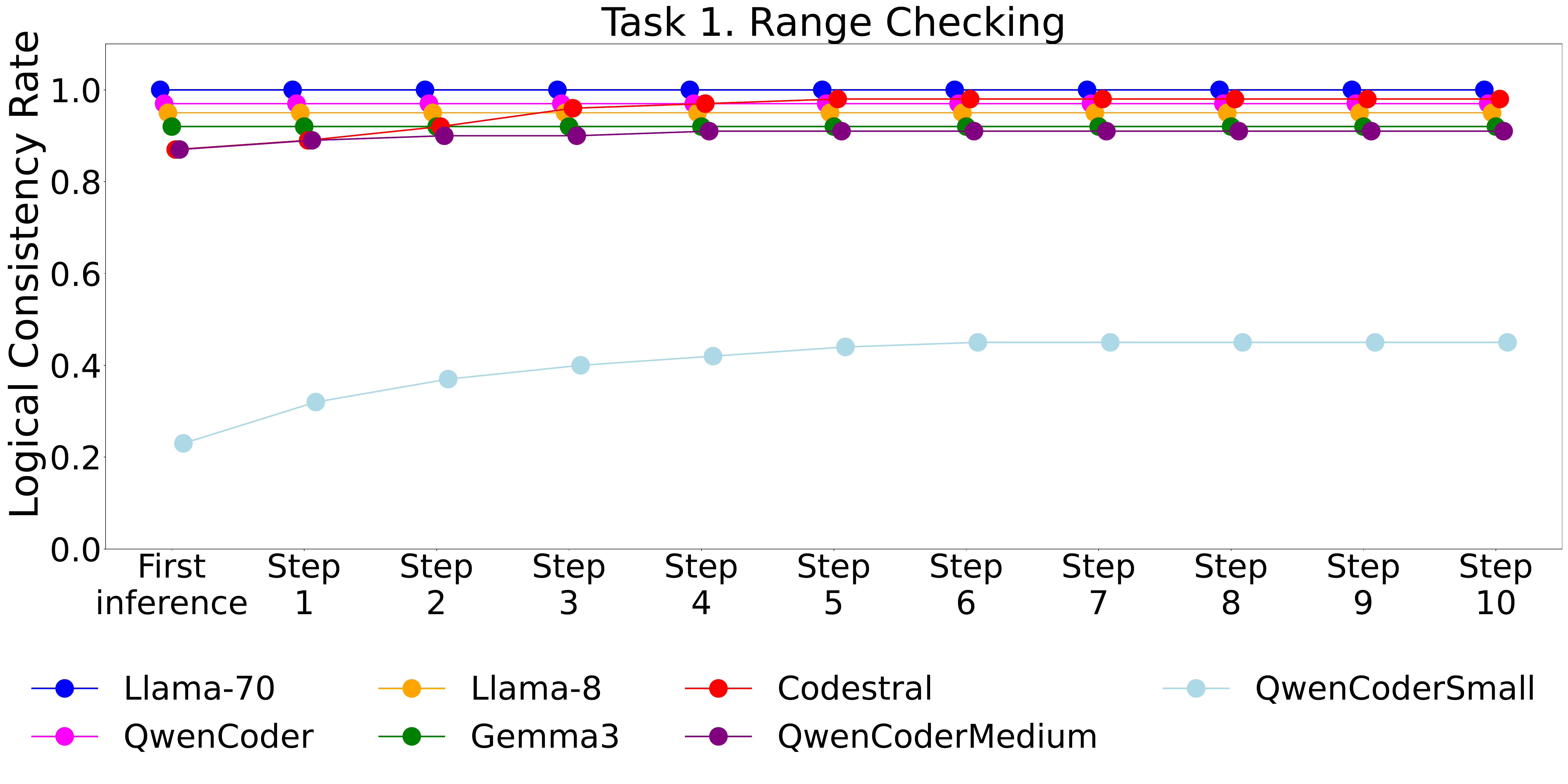} 
  \caption {Task 1 Logical Consistency Rate (LCR) for first inference and for each of the refinement steps.}
  \label{fig:task1_OA_before_after_steps_comp}
\end{figure}
\begin{figure}[t]
    \centering
  \includegraphics[width=\columnwidth]{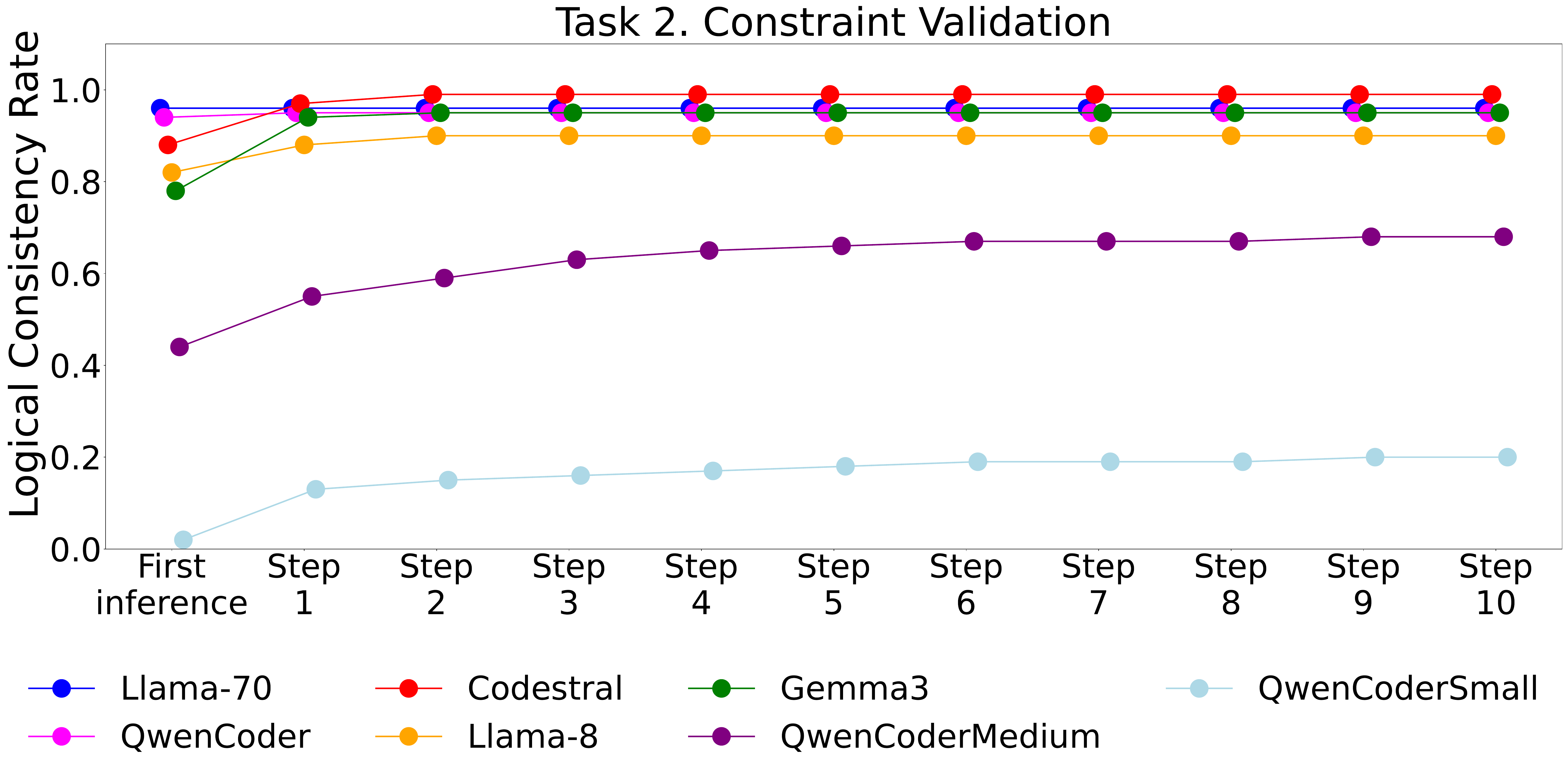} 
  \caption {Task 2 Logical Consistency Rate (LCR) for first inference and for each of the refinement steps.}
  \label{fig:task2_OA_before_after_steps_comp}
\end{figure}

Key numerical results are summarized in Table \ref{tab:tasks_results}, while Figure \ref{fig:task1_OA_before_after_steps_comp} and \ref{fig:task2_OA_before_after_steps_comp} shows changes in LCR in successive stages of iterative refinement for both tasks. Figures \ref{fig:task1_OA_before_after_comp}, \ref{fig:task2_OA_before_after_comp} shows the same changes but with detailed results for each parameter. 
Figure \ref{fig:lcr_vs_models} shows the distribution of LCR metric values depending on the size of the model.

\begin{table}[h]
  \centering
  \resizebox{0.5\textwidth}{!}{
  \small
  \begin{tabular}{cccccccc}
    \hline
    \multicolumn{8}{c}{Range Checking evaluation results} \\
    \multicolumn{1}{c}{Model} & \multicolumn{2}{c}{First inference} & \multicolumn{2}{c}{After all refinements steps} & \multicolumn{2}{c}{Improvement} & \multicolumn{1}{c}{Number} \\
      & FSR & LCR & FSR & LCR R & $\Delta$ FSR & $\Delta$ LCR & of iterations \\
    \hline
    Llama-70 & \textbf{1.0} & \textbf{1.00} & \textbf{1.0} &\textbf{ 1.0} & 0.00 & 0.00 & 0 \\
    Codestral & 1.0 & 0.87 & 1.0 & 0.98 & 0.00 & 0.11 & 5 \\
    QwenCoder & 1.0 & 0.97 & 1.0 & 0.97 & 0.00 & 0.00 & 0 \\
    Gemma3 & 1.0 & 0.92 & 1.0 & 0.92 & 0.00 & 0.00 & 0 \\
    Llama-8 & 1.0 & 0.95 & 1.0 & 0.95 & 0.00 & 0.00 & 0 \\
    QwenCoderMedium & 1.0 & 0.87 & 1.0 & 0.91 & 0.00 & 0.04 & 4 \\
    QwenCoderSmall & 1.0 & 0.23 & 1.0 & 0.45 & 0.00 & \textbf{0.22} & 6 \\
    \hline
  \end{tabular}
  }
\end{table}

\begin{table}[h]
  \centering
  \resizebox{\columnwidth}{!}{
  \small
  \begin{tabular}{cccccccc}
    \hline
    \multicolumn{8}{c}{Constraint Validation evaluation results} \\
    \multicolumn{1}{c}{Model} & \multicolumn{2}{c}{First inference} & \multicolumn{2}{c}{After all refinements steps} & \multicolumn{2}{c}{Improvement} & \multicolumn{1}{c}{Number} \\
      & FSR & LCR & FSR & LCR R & $\Delta$ FSR & $\Delta$ LCR & of iterations \\
    \hline
    Llama-70 & \textbf{1.0} & \textbf{0.96} & 1.0 & 0.96 & 0.00 & 0.00 & 0 \\
    Codestral & 1.0 & 0.88 & \textbf{1.0} & \textbf{0.99} & 0.00 & 0.11 & 2 \\
    QwenCoder & 1.0 & 0.94 & 1.0 & 0.95 & 0.00 & 0.01 & 1 \\
    Gemma3 & 1.0 & 0.78 & 1.0 & 0.95 & 0.00 & 0.17 & 2 \\
    Llama-8 & 0.99 & 0.82 & 1.0 & 0.90 & 0.00 & 0.08 & 2 \\
    QwenCoderMedium & 1.0 & 0.44 & 1.0 & 0.67 & 0.00 & \textbf{0.23} & 6 \\
    QwenCoderSmall & 1.0 & 0.02 & 1.0 & 0.19 & 0.00 & 0.17 & 6 \\
    \hline
  \end{tabular}
  }
  \caption{Formalization Success Rates (FSR) and Logical Consistency Rate (LCR) in Range Checking (top) and Constraint Validation (bottom) tasks. $\Delta$ corresponds to the change between the \textit{first inference} and after all \textit{refinement steps} that made a difference.}
  \label{tab:tasks_results}
\end{table}

\section{Related work}
Recent advancements in large language models (LLMs) have significantly enhanced capabilities in logical reasoning and code generation. However, limitations remain in reliably integrating expert knowledge and ensuring logical consistency.

In the domain of logical reasoning and verification, early works began exploring the integration of external reasoning tools with LLMs. For instance, McGinness and Baumgartner \cite{mcginness2024automated} combined Automated Theorem Provers with LLMs, improving logical accuracy via a neuro-symbolic method tested on the PRONTOQA benchmark. Around the same time, LeanReasoner \cite{jiang2024leanreasoner} and Logic-LM \cite{pan2023logic} formalized reasoning tasks as theorems to be externally solved, while LINC \cite{olausson2023linc} and Symbolic Chain-of-Thought \cite{xu2024faithful} translated natural language into first-order logic to facilitate symbolic deduction.

As the field progressed, researchers began to stress-test and scaffold reasoning behaviors. Wan et al. \cite{wan2024logicasker} proposed logic scaffolding to expose inferential weaknesses in LLMs, and Zhou et al. \cite{zhou2024rulearena} introduced RULEARENA to evaluate rule-guided reasoning in multi-rule environments. Meanwhile, Hong et al. \cite{hong2023closer} highlighted structural fallacies in LLM outputs using the FALLACIES dataset, revealing core weaknesses in self-verification. Addressing these, Liang et al. \cite{liang2024improving} introduced a fine-grained “Step CoT Check” technique for better detection and correction of reasoning errors, while Wan et al. \cite{wan2024cot} developed the CoT Rerailer framework, which integrates both error detection and correction to boost reasoning fidelity.

In pursuit of scalability and automation, Luo et al. \cite{luo2024improve} presented OmegaPRM, a Monte Carlo Tree Search-based method that significantly improved performance on mathematical benchmarks like MATH500 and GSM8K. Liang et al. \cite{liang2024improving} later extended this line of work by proposing a collaborative verification framework that distributes inference computations to further improve scalability and reasoning robustness.

Complementary efforts have injected formal logical structures directly into the LLMs’ reasoning processes. For example, Logic-of-Thought and Logic Agent \cite{liu2024logic} incorporate explicit logical rules to ensure more valid outputs. Several symbolic reasoning-focused studies also emerged around this time \cite{wysocka2024syllobio, dalal-etal-2024-inference, quan-etal-2024-verification, quan-etal-2024-enhancing, meadows-etal-2024-symbolic, meadows2025controllingequationalreasoninglarge}, some bridging with general code generation tasks \cite{wang2023review, yadav2024pythonsaga}.

Our investigation builds upon these developments, especially research into iterative self-correction and refinement. Feedback-loop frameworks such as LLM-ARC \cite{kalyanpur2024llm} and Determlr \cite{sun2024determlr} show promise in improving logical outputs. However, our findings indicate that, even when syntactic outputs are correct, logical inconsistencies in the translation of expert rules often persist—echoing issues highlighted in prior work on code verification and logical consistency \cite{wu2023lemur, feng2023language, feng2024language, chen2022program}. To address this, other studies have proposed cross-model correction \cite{li2024leveraging} and more robust feedback mechanisms \cite{yang2024enhancing}.

With regard to stability and evaluation, Liu et al. \cite{liu2024your} introduced G-Pass@k and LiveMathBench to capture discrepancies between theoretical potential and operational stability. Lifshitz et al. \cite{lifshitz2025multi} advanced this further through Multi-Agent Verification, which employs multiple verifiers at inference time to boost accuracy and scalability. Confidence aggregation techniques such as CER, introduced by Razghandi et al. \cite{razghandi2025cer}, also proved effective in enhancing reliability across diverse reasoning tasks.

Verification methods have grown more sophisticated. PROVE \cite{toh2024not} applies program-based checks to improve consistency in mathematical reasoning, outperforming majority-voting baselines. Similarly, Min et al.’s IdentityChain framework measures and strengthens consistency between specifications and generated outputs, emphasizing the importance of verification in code synthesis.

In the area of code generation and efficiency, researchers have tackled logical coherence and hallucination. Tian et al. \cite{tian2025codehalu} addressed this directly with the CodeHalu algorithm and the CodeHaluEval benchmark, which highlight reliability gaps. Li et al. \cite{li2024model} proposed GRACE, a model-editing approach that enables precise corrections without degrading unrelated capabilities. In parallel, ECCO \cite{waghjale2024ecco} introduced a benchmark to assess trade-offs between correctness and computational efficiency, advocating for execution-based feedback.

Finally, in formal proof generation, Lin et al. \cite{lin2025goedel} introduced Goedel-Prover, which demonstrated superior performance in automated theorem proving, outperforming models like DeepSeek-Prover-V1.5. Leang et al. \cite{leang2025theorem} complemented this with TP-as-a-Judge, which uses theorem provers to validate intermediate reasoning steps, thereby increasing model precision in mathematical benchmarks.

Collectively, these works illustrate a rich and evolving landscape. Despite major strides in integrating symbolic reasoning, verifying outputs, and improving generalization in code generation, substantial gaps remain—particularly in ensuring that expert rules are faithfully and logically represented in LLM outputs.

\section{Conclusions}

We introduced ExKLoP, a framework designed to evaluate LLMs’ ability to integrate expert knowledge into logical reasoning and self-correction. Leveraging an extensible dataset of engineering premises and validation points, our approach systematically assesses both the syntactic fluency and logical correctness of Python code for critical tasks like range checking and constraint validation.

Our experiments demonstrated the high efficiency of most tested models in generating syntactically and logically correct Python code. Coding-oriented models such as Codestral-22B and QwenCoder2.5-14B performed well across both tasks, achieving final Logical Correctness Rates (LCRs) of 98\% and 97\% in range-checking, and 99\% and 95\% in constraint validation, respectively. However, Codestral-22B initially showed some anomalies, performing below smaller models with an LCR of 88\% - 4 to 9 percentage points lower than its counterparts. Smaller general-purpose models like Llama3-8B and Gemma3-12B also delivered satisfactory results, with LCRs of 95\% and 92\% in Task 1, and 90\% and 95\% in Task 2, respectively, showing performance not far behind the specialized coding models. QwenCoder2.5-3B performed adequately in the range-checking task with a 91\% LCR but experienced a significant drop in the constraint validation task, ending with 67\%. The smallest model, QwenCoder2.5-1.5B, performed poorly in both tasks, achieving only 45\% in range-checking and 19\% in constraint validation. Self-correction proved to be a key contributor to performance improvements across all models. Our experiments revealed that the number of iterations varied by model, with each reaching a stopping point no later than the 6th iteration.

Overall, ExKLoP offers a robust evaluation platform that streamlines the selection of effective models for self-correcting systems while clearly delineating error types. This framework establishes a valuable benchmark and lays the groundwork for future research aimed at enhancing logical consistency and the integration of expert knowledge in AI-based Expert systems.



\section{Limitations}
Our study demonstrates the potential of LLMs for expert knowledge integration and constraint validation, but few limitations must be acknowledged:
\newline
\textbf{Dataset Constraints:} Our experiments were conducted on a domain-specific dataset (vehicle operation parameters). This limits the generalizability of our findings to other fields, such as biomedicine or finance. The dataset includes synthetic premises and validation points. 
\newline
\textbf{Prompting Limitations:} Our approach relied on In-Context Learning (ICL) without advanced prompting strategies like Chain-of-Thought (CoT) reasoning. These techniques might improve logical accuracy. The iterative refinement process improved performance but still relied on model-generated self-corrections, which may introduce biases or reinforce incorrect patterns.
\newline






\bibliography{mybibfile}

\end{document}